\documentclass[twocolumn]{article}

\usepackage{times}
\usepackage{epsfig}
\usepackage{graphicx}
\usepackage{amsmath}
\usepackage{amssymb}

\usepackage{color}
\usepackage{multirow}
\usepackage{amsfonts}
\definecolor{cardinal}{rgb}{0.77, 0.12, 0.23}
\definecolor{officegreen}{rgb}{0.0, 0.5, 0.0}
	\definecolor{lightbrown}{rgb}{0.71, 0.4, 0.11}
\definecolor{limegreen}{rgb}{0.2, 0.8, 0.2}

\usepackage{adjustbox}

\usepackage[ruled,lined]{algorithm2e}

\title{Interpretable Few-shot Learning with Online Attribute Selection}
\author{
  Mohammad Reza Zarei\\
  School of Computer Science\\
  Carleton University\\
  Ottawa, Canada\\
  \texttt{Mohammadrezazarei@cmail.carleton.ca}
  \and
  Majid Komeili\\
  School of Computer Science\\
  Carleton University\\
  Ottawa, Canada\\
  \texttt{Majid.Komeili@carleton.ca}
}
\date{}

\begin{document}

\maketitle

\begin{abstract}
Few-shot learning (FSL) presents a challenging learning problem in which only a few samples are available for each class. Decision interpretation is more important in few-shot classification due to a greater chance of error compared to traditional classification. However, the majority of the previous FSL methods are black-box models. In this paper, we propose an inherently interpretable model for FSL based on human-friendly attributes. Previously, human-friendly attributes have been utilized to train models with the potential for human interaction and interpretability. However, such approaches are not directly extendible to the few-shot classification scenario. Moreover, we propose an online attribute selection mechanism to effectively filter out irrelevant attributes in each episode. The attribute selection mechanism improves accuracy and helps with interpretability by reducing the number of attributes that participate in each episode. We further propose a mechanism that automatically detects the episodes where the pool of available human-friendly attributes is insufficient, and subsequently augments it by engaging some learned unknown attributes. We demonstrate that the proposed method achieves results on par with black-box few-shot learning models on four widely used datasets. We also empirically evaluate the level of decision alignment between different models and human understanding and show that our model outperforms the comparison methods based on this criterion.
\end{abstract}

\section{Introduction}
\label{intro}
In recent years, deep learning has been able to achieve impressive success in computer vision tasks, including object recognition. This success heavily relies on the availability of abundant labeled examples. However, accessing sufficient data is not always possible due to the limitations and challenges in data collection and annotation. Therefore, a new learning paradigm known as few-shot learning (FSL) has been considered in which the model has to generalize to novel classes with just a few training instances \cite{9897572}. 

FSL models are first trained on a set of base classes with abundant samples, preparing to classify novel classes with only a few examples, where the base and novel classes are entirely disjoint. This models human generalization ability that can effectively transfer the insight obtained from prior knowledge to future tasks with limited supervision \cite{mangla2020charting}. Each standard few-shot classification task is called an episode. In each episode, N distinct classes are sampled from novel classes. For each of these classes, K instances are sampled to form a labeled training set, known as support set and an unlabeled query set which the model has to classify correctly \cite{Sung_2018_CVPR}. This setting is known as N-way K-shot. 


Recently, few-shot learning models have been able to achieve remarkable performance. However, they usually neglect decision interpretability. As their decision can not be interpreted by human, they are considered to be black-box models. In recent years, using black-box models in real world has drawn growing concerns. Since few-shot learning is an intrinsic challenging task with a higher chance of error compared to traditional classification, decision interpretation is more critical in this context. Decision interpretability can be achieved by performing post-hoc analysis approaches after training a model. However, these methods add an additional step to the process. Moreover, choosing the most appropriate post-hoc interpretability approach is a challenge. Alternatively, we can design implicitly interpretable models so that their decisions can be interpreted during inference without requiring any additional steps. Since decision interpretation is usually easier in these models compared to applying post-hoc models, non-expert users may prefer them.


Recently, semantic attributes have been used as a valuable source for training implicitly interpretable models. Concept Bottleneck Models (CBMs) \cite{koh2020concept} are the main approaches in this direction that focus on aligning the intermediate layer of a neural network with attributes. These models use all available attributes for every classification decision. However, as we will show in Section \ref{eval}, inclusion of irrelevant attributes may harm the accuracy and decision interpretabilty. Moreover, the base structure of these models is not directly applicable to the few-shot setting.

In this paper, we follow the idea of CBMs and propose an inherently interpretable FSL model based on human-friendly attributes. At the core of the proposed method is interpretability and the ability to trace decisions back to human-friendly attributes. We believe that in many real-world scenarios, if a task can be solved along the human-friendly dimensions (attributes), then that should be the preferred way compared to black-box approaches, and equally important is to have a mechanism to detect and deal with novel tasks where the available attributes are not adequate. We assume that there exists a large pool of attributes in the same way that we assume there exists a large pool of tasks (base classes) for FSL. However, the challenges are 1) how to effectively use attributes to do novel tasks in an interpretable way 2) how to know when a novel task cannot be properly done using the available pool of attributes and 3) what to do in such circumstances. We have proposed a solution for each of the three issues. To the best of our knowledge, our work is the first to address these issues in a unified framework.

\begin{figure}
  \centering 
  \includegraphics[width=0.9\linewidth]{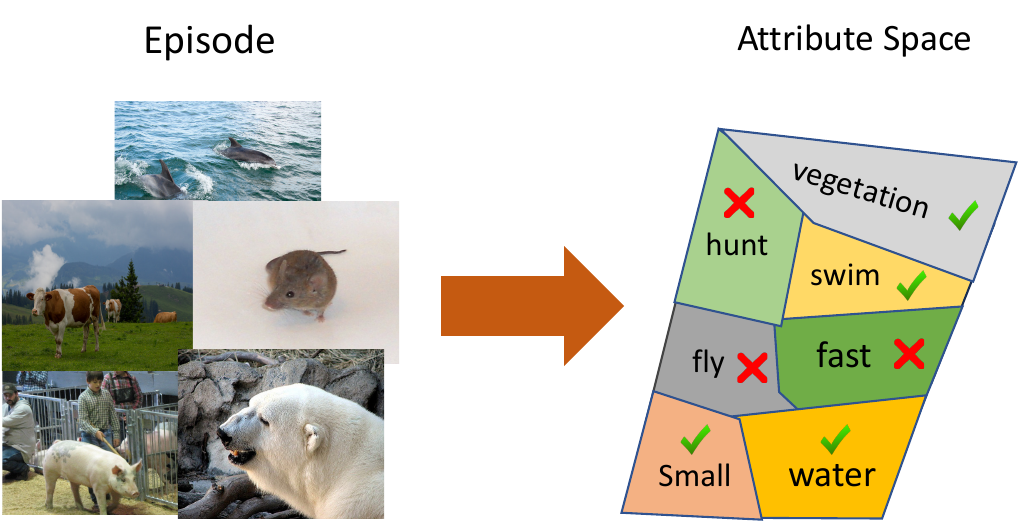}
  \caption{In each few-shot episode, we filter out irrelevant attributes by proposing an online attribute selection mechanism. This improves interpretability and few-shot classification accuracy.}
  \label{fig.motivation}
\end{figure}

The contributions to this paper are as follows:

1) We propose a mechanism for FSL based on human-friendly attributes. It does not require attribute labels during inference.
2) Not all attributes may be relevant to the task in each episode of few-shot learning. Therefore, we propose an online attribute selection mechanism that enhances interpretability and few-shot classification accuracy by filtering out irrelevant attributes in each episode. This is the main contribution of our paper and the high-level motivation is shown in Figure \ref{fig.motivation}.
3) We propose a mechanism that detects episodes where the pool of known human-friendly attributes is inadequate for the novel task and subsequently augments it by engaging some unknown attributes. These unknown attributes are learned to complement the human-friendly attributes without overlapping them. This is achieved by minimizing mutual information between unknown attributes and the human-friendly attributes during training. Such overlap is undesirable because, in the presence of a large overlap, altering only a known attribute during the test-time intervention may not simulate its actual impact on the prediction. 

\begin{figure*}
  \centering 
  \includegraphics[width=0.9\linewidth]{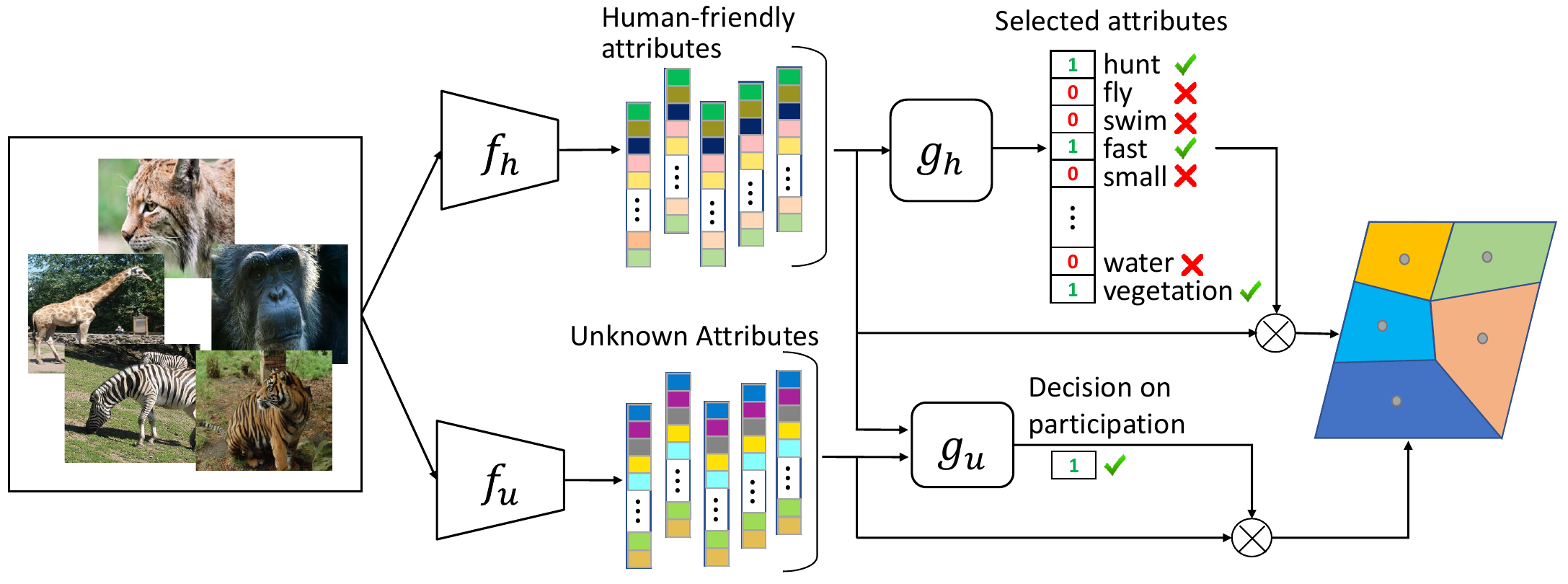}
  \caption{Few-shot classification using our framework. Top: The process of predicting and selecting relevant human-friendly attributes for the current episode. Bottom: The process of predicting unknown attributes and deciding whether the unknown attributes should participate in the current episode or not.}
  \label{fig.diagram}
\end{figure*}

Figure \ref{fig.diagram} shows the process of learning and selecting relevant attributes in an episode of 5-way 1-shot setting.
Black-box FSL models are more free to concentrate solely on attaining high accuracy. Despite this, the proposed interpretable method performs on par with four previously state-of-the-art black-box FSL models on four widely used datasets in terms of accuracy and also achieves the highest decision alignment with human understanding compared to all compared methods.

\section{Related Work}
Few-shot learning approaches can be roughly grouped into two main categories, optimization-based methods \cite{rusu2018metalearning,antoniou2018how} and metric-based approaches \cite{Sung_2018_CVPR,Proto}. Optimization-based approaches aim at learning a base network that can be adapted to a target task in a few gradient steps \cite{cao2021concept}. On the other hand, metric-based approaches attempt to learn a unified metric space across different tasks in which each task could be solved using a similarity or distance measure or by using a simple classifier. In this category, MatchingNet \cite{vinyals2016matching} utilizes a memory module to effectively leverage the information in each task and performs one-shot classification using cosine metric. ProtoNet \cite{Proto} learns an embedding space in which the samples of each class could be clustered around the corresponding class-specific prototype. RelationNet \cite{sung2018learning} attempt to learn a network to measure the similarity between few-shot pairs. 
The proposed framework lies in the second category and follows ProtoNet. However, unlike this method and all of the above-mentioned approaches that lack interpretability, the decision of our framework can be interpreted by human since it targets attribute space as a human-friendly metric space.


Research in few-shot learning with interpretability in mind has been scarce. Several prior work has attempted to achieve post-hoc interpretability by improving feature learning. CPDE \cite{zou2020compositional} utilizes a self-supervision loss based on the split-order prediction, in addition to classification loss, to discover part-related primitives. This approach alleviates the semantic gap and leads to an improvement in post-hoc interpretability. Various methods have incorporated attention module to enhance post-hoc interpretability. RENet \cite{kang2021relational} employs self-correlation module to extract structural patterns and a cross-correlation module to produce co-attention between images. SARN \cite{8794911} proposes a self-attention relation network that learns non-local information and improves extracted features. 
Different from the above-mentioned approaches, COMET \cite{cao2021concept} attempts to propose an intrinsic interpretable FSL model by learning along human interpretable concepts. 
However, COMET has a number of limitation: it requires concept annotations at pixel-level. It requires such fine-grained annotations not only for base classes but also for test samples. Moreover, it requires all the concepts to be present in each sample and shared among all images.
In contrast, our method does not require concept annotations at test time. It requires only concept annotations in the form of binary attributes. Annotations are required only for training classes also known as base classes. Moreover, each attribute can be present or absent in some images. Therefore, it makes our framework applicable to a wider range of scenarios.

\textit{Relation to Concept Bottleneck Models:}
%
Recently, semantic attributes have been used to learn models that bring the possibility of human interaction and achieve interpretability. Concept Bottleneck Models (CBMs) \cite{koh2020concept} are the main approaches in this direction that aligns an intermediate layer of a neural network with attributes which subsequently are used to predict the output.
The proposed method follows the idea of concept bottleneck models with a focus on extending them to few-shot classification setting since base structure is not compatible with FSL. The conventional CBMs consider classifier weights on top of the attribute layer to map the attributes to each class by learning proper weights for them. However, this approach is not directly applicable to FSL since the base and novel classes are disjoint and the classifier weights learned on base classes will not be useful for novel classes. A trivial solution for this problem is to learn classifier weights specifically for each episode on novel classes. However, the limited number of samples for each class leads to overfitting and insufficient performance. Moreover, we argue that an interpretation based on all the available attributes in CBMs may not be advantageous particularly if the number of attributes are high. Also, not all of the attributes are relevant to the task at each episode. Hence, we propose an attribute selection mechanism to filter out irrelevant attributes in each episode and perform few-shot classification in a non-parametric setting.

\textit{Relation to Zero-shot Learning:}
Zero-Shot Learning (ZSL) utilizes semantic information to compensate unavailability of training instances for unseen classes and serve a bridge to transfer the knowledge from seen classes to unseen ones \cite{ZAREI2021801}. 
Similar to zero-shot learning, we also leverage human-friendly attributes. However, in ZSL attributes are mainly used to account for the lack of training samples in novel classes, rather than interpretibility. Moreover, zero-shot learning uses the attributes for both its training and testing classes to achieve knowledge transferability, but we use attributes just for training classes known as base classes in few-shot learning problem.

\textit{Interpretability through attribute localization:} Attribute localization in images has recently gained attention, particularly in zero-shot learning (ZSL), as a means to bridge the semantic gap between domains and achieve interpretability. Based on the type of interpretability achieved through attribute localization, models can be categorized as either post-hoc explainable or intrinsically interpretable.
In the post-hoc explainable category, GEM-ZSL \cite{Liu_2021_CVPR} introduces a gaze estimation module that employs predefined human-friendly attributes to guide localized attribute learning, thus enhancing encoder training. Similarly, the Attribute Prototype Network (APN) \cite{Xu2022} improves global image features by learning local attributes, thereby enhancing representation. However, while these methods leverage attribute localization to strengthen representation, the localized features are not directly utilized in decision-making, which serves to improve the post-hoc explainability of the models.
In the realm of intrinsic interpretability, DIML \cite{10334048} decomposes the similarity between image pairs into part-wise contributions using optimal transport theory. Nevertheless, this measured similarity is confined to unknown parts, and it does not indicate the specific points or aspects responsible for the similarity. Several works that utilize attribute localization have explored the learning of representations in a human-friendly attribute space derived from images, as well as performing zero-shot classification in that space. This approach results in intrinsically interpretable models where decisions can be based on various attributes. For instance, TransZero \cite{chen2022transzero} constructs attribute-guided Transformers that generate locality-augments features by mapping attributes to specific image regions for classification. Additionally, RSAN \cite{10.1145/3459637.3482471} develops a fine-grained recognition branch that extracts each attribute from specific regions of an image and utilizes these attributes for recognition. The image encoder in this branch is regularized through attribute regression with semantic knowledge, allowing for the extraction of more robust and attribute-related visual features. RIAE \cite{HU2022984} learns local attribute features and leverages them to obtain global attribute features for zero-shot classification by utilizing a region graph network and an attribute feature embedding. 
Our model also qualifies as an intrinsically interpretable model; however, it specifically addresses the few-shot learning problem by employing an online attribute selection mechanism. Unlike the discussed approaches that primarily focus on enhancing localization and attribute mapping, our model enhances interpretability and decision-making efficiency by filtering and aggregating a limited subset of human-friendly attributes for each episode. This approach enables clearer decision rationales.

\textit{Using attribute supervision in Few-shot Learning:} Semantic attributes have also been utilized in Few-shot learning. In \cite{tokmakov2019learning}, Tokmakov and et. al employed attributes to improve the compositionality of image representation. Auto-ACNet \cite{zhang2021auto} incorporates attributes within its framework to offer additional supervision for the representation learning process and to mitigate the domain shift problem. This work employs the neural architecture search technique DARTS \cite{liu2019darts} to identify the superior network for few-shot learning task. Dual TriNet \cite{chen2019multi} learns an autoencoder framework and leverages semantic attributes to augment visual features for few-shot classification. AGAM \cite{huang2021attributes} learns an attribute-guided attention module to learn more discriminative features by focusing on important channels and regions. Although these methods also leverage semantic information similar to our method, they have used attributes mainly to augment the visual information to achieve higher classification accuracy without any interpretability consideration whatsoever. Although post-hoc saliency  approaches can be applied to any of these models, these explanation methods are exposed to the risk of providing unreliable interpretation \cite{adebayo2018sanity}. This is in contrast to our proposed method where interpretability is at core and the attributes act as bottleneck to ensure decisions can be traced back to human-friendly attributes.

\section{Proposed Method}

Suppose that we are given a labeled dataset $\mathcal{D}^{b}  = {\{(x_i,y_i)\}}_{i=1}^{N_b}$ for base classes $\mathcal{Y}_b$ with sufficient images for each class, and a limited number of labeled samples for novel classes $\mathcal{Y}_n$ known as support set $\mathcal{S}_n={\{(x_i,y_i)\}}_{i=1}^{N_s}$ where ${\mathcal{Y}_b} \cap {\mathcal{Y}_n} = {\emptyset}$. The goal is to predict the label of a query set $\mathcal{Q}_n = {\{(x_i,y_i)\}}_{i=1}^{N_q}$ which also belongs to novel classes ${\mathcal{Y}_n}$. It is assumed that each image $x_i$ in $\mathcal{D}^{b}$ is associated with a binary attribute vector $a_{x_i} \in \mathbb{R}^A$, where $A$ denotes the number of attributes,  
 specifying the properties related to the object in the image. These attributes help achieve interpretability.

Our interpretable attribute-based model consists of an attribute predictor $f_{h}$ that attempts to map each image to corresponding human-friendly attribute vector and attribute selector $g_{h}$ that selects the most relevant human-friendly attributes for each episode of few-shot classification. To further improve accuracy, we enable leveraging additional attributes (unknown attributes) by training an unknown attribute predictor $f_u$, and an unknown attribute participation detector $g_u$ that decides whether unknown attributes should participate in the current episode or not. The process is shown in Figure \ref{fig.diagram} for a 5-way 1-shot episode. The proposed framework is trained on the images of base classes. Then, it is used to perform few shot classification on novel classes in target space. 

\subsection{Attribute Predictor Network}

In this subsection, we explain the human-friendly attribute predictor network $f_{h}$. This network acts as a feature extractor for few-shot classification. The conventional feature extractors designed for few-shot classification usually just focus on learning some features to achieve a high accuracy, neglecting the interpretability of the learned features and the decision made by using those features. Therefore, we target the attribute space as a meaningful embedding space for few-shot classification and train $f_{h}$ to predict attributes for each image. 

The attribute predictor network $f_{h}$ attempts to map each image ${x_i}$ to corresponding attribute vector $a_{x_i}$ by predicting $\hat{a}_{x_i} \in \mathbb{R}^A$:
\begin{equation}
    \label{eq__1}
    \hat{a}_{x_i} = f_{h}(x_i)
\end{equation}

We define the loss function for attribute prediction as a sample-wise weighted binary cross entropy between the actual attribute values and the predicted values: 
\begin{equation}
    \label{eq__2}
    l_{att} = -\sum_{x_i \in \mathcal{D}^{b}}\sum_{j=1}^{A}w_{x_i}^{j}[a_{x_i}^{j}\log {\hat{a}_{x_i}^j} + (1-a_{x_i}^{j})\log {(1-\hat{a}_{x_i}^j)]}
\end{equation}
where $\hat{a}_{x_i}^j$ and $a_{x_i}^{j}$ represent the predicted and actual values of the $j^{th}$ attribute of image $x_i$, respectively. Additionally, $w_{x_i}^{j}$ is the weight associated with the loss imposed by the $j^{th}$ attribute of image $x_i$ and is equal to $1/n_{x_i}^{j}$ where $n_{x_i}^{j} = |\{a_{x_i}^{k} \in a_{x_i}|a_{x_i}^{k} = a_{x_i}^{j}\}|$. Intuitively, $n_{x_i}^{j}$ is the number of attributes in ${x_i}$ with actual value equal to  $a_{x_i}^{j}$. 
Since the majority of the attributes are usually absent (equal to 0) in each image, utilizing standard binary cross entropy results in a bias toward predicting the value of the attributes as 0. Therefore, instead of considering equal weights for different attributes of each sample, the loss value for each present attribute is normalized by the total number attributes that are present in that sample. Likewise, the loss related to the absent attributes are normalized by the total number of absent attributes in that sample. By considering this weighted binary cross entropy, the weight of the attributes with minority value in each sample is increased.

After training on the images of base classes using $l_{att}$ in Equation \ref{eq__2}, the attribute predictor network can be used to predict the attributes of novel images and perform few-shot classification in attribute space. A class-specific prototype $\hat{c} \in \mathbb{R}^A$ is calculated in the attribute space for each class $y$ by averaging the attributes of the support set:
\begin{equation}
    \label{eq__3}
    \hat{c}_y = \frac{1}{|S_y|} \sum \limits_{x_i \in {S_y}} \hat{a}_{x_i}
\end{equation}
where $|S_y|$ is the number of images in the support set of the class $y$. To specify the class of a query image $x_q$, the distance of its predicted attributes $\hat{a}_{x_q}$ is calculated from the class-specific prototypes. This distance is used to measure the probability of the query image belonging to each class. Therefore, the probability of assigning the query image $x_q$ to the class $y$ is estimated as:
\begin{equation}
    \label{eq__4}
    p(y|x_q) = 
    \frac{exp(-d(\hat{a}_{x_q},\hat{c}_y))}
    {\sum_{y' \in {\mathcal{Y}_s}}exp(-d(\hat{a}_{x_q},\hat{c}_{y'}))}
\end{equation}
where $d(.)$ is a distance function.

Although the probability of assigning an image to one class can be interpreted in terms of the difference between the predicted attributes of query image and the prototypes of novel classes, in fact this interpretation may not be advantageous since the number of attributes is usually high. Moreover, some attributes may not be relevant to the classes of individual episodes of few-shot classification. Therefore, employing all attributes may degrade the classification accuracy.

To tackle the above problems, an attribute selector network is proposed in the next sub-section that will be employed to select relevant attributes in each episode of few-shot recognition.






\subsection{Attribute Selector Network}

After training the attribute predictor network $f_{h}$, the weights of $f_{h}$ are frozen and a selector network $g_{h}$ is trained to select useful attributes for few-shot classification. To mimic few-shot recognition tasks during training of this network, the images of base classes are sampled in the form of episodes. In each episode, N classes are randomly selected from the base classes. For each of these classes, a support set $\mathcal{S}_b$ consists of K instances and another subset serving as the query set $\mathcal{Q}_b$ are sampled.

The attribute predictor network exploits the attributes of the support set to select relevant attributes for classification of the query set. The choice of whether the $i^{th}$ attribute is selected is modeled by a categorical distribution with two possible outcomes, 1 or 0. This distribution deciding on participation of the $i^{th}$ attribute is represented with:

\begin{equation}
    p(s^{(i)}) = 
    \begin{cases}
        \pi_i & \text{if } s^{(i)} = 1, \\
        1 - \pi_i & \text{if } s^{(i)} = 0.
    \end{cases}
\end{equation}
The number of categorical distributions will be equal to the number of attributes. In each episode, by sampling from these categorical distributions, the binary state vector $s \in \mathbb{R}^A$ is obtained, determining which attributes should be used in that episode.

In each episode, the attribute vectors for the support set of each class are averaged to obtain the prototype of each class as in Equation \ref{eq__3}. We consider these prototypes as a sequence of inputs and feed them to $g_h$ which consists of a Bi-LSTM followed by a feed-forward network, denoted by $g_{h\_\text{LSTM}}$ and $g_{h\_\text{MLP}}$. Bi-LSTM consists of two independent LSTMs, one processing the input sequence in a forward manner and the other in a backward manner. The Bi-LSTM effectively aggregates the information from the attribute prototypes of all classes and provides unified features for them. Since the set of prototypes is not inherently ordered and does not follow a sequence, it may be thought that changing their order as input to the Bi-LSTM could affect the performance of the model. However, as the training of the network is performed episodically and the classes are randomly sampled with different orders during training, the sensitivity of the model's performance to different permutations of the input to the Bi-LSTM will be minimized. We further investigate the impact of different permutations of Bi-LSTM input on the model's performance in \ref{app_lstm}.

The prototypes of all classes $\hat{C} \in \mathbb{R}^{N\times A}$ are passed to the Bi-LSTM $g_{h\_\text{LSTM}}$ and the last hidden state of the forward and backward LSTMs are retrieved from it:

\begin{equation}
    (h_N,{h'_N}) = g_{h\_\text{LSTM}}(\hat{C})
\end{equation}
where $h_N$ and $h'_N$ are the last hidden states of the forward and backward LSTMS, respectively. These two vectors are then concatenated and passed through the feed-forward network  $g_{h\_\text{MLP}}$ with sigmoid activation function in the final layer to produce the probabilities of the categorical distributions. The output is a vector with a dimension equal to the number of attributes $A$ where the $i^{th}$ element denoted as $\pi_i$, represents the probability of the $i^{th}$ attribute being selected. Clearly, the probability of the same attribute being disregarded is $1-\pi_i$. 


While the output for the binary-trial of each categorical distribution can be determined by sampling, it is a non-differentiable operation. To tackle this problem, we use Gumbel Softmax estimator \cite{Gumbel, maddison2017the} that enables a differentiable sampling process from categorical distributions using reparameterization trick \cite{Gumbel3,pmlr-v32-rezende14}. Therefore, the output value that specifies whether the $i^{th}$ attribute is selected or not is determined as:
\begin{equation}
\label{eq__5}
    s^{(i)} = \\ \frac{\exp{((\log \pi_i + \gamma_i)/\tau)}}{\exp{((\log \pi_i + \gamma_i)/\tau)} + \exp{((\log \pi_{i^\prime} + \gamma_{i^\prime})/\tau)}}
\end{equation}
where $\pi_{i^\prime}$ = $1-\pi_i$, $s \in \mathbb{R}^A$,  $\gamma_i$ and $\gamma_{i^\prime}$ are two Gumbel noises sampled from standard Gumbel distribution. Furthermore, $\tau$ is a temperature parameter determining the degree of discreteness. When $\tau \rightarrow 0$, the output becomes discrete. During training, we use Gumbel Softmax estimator to specify the states of embedding spaces (i.e. selected attributes). During inference, we simply select the binary value with higher probability as the state of each embedding space (i.e. no Gumbel Softmax sampling). 

The state vector $s$ is used as the weight vector to scale the dimensions of the attribute space. Therefore, following Equation \ref{eq__4}, the probability of assigning the query image $x_q$
to the class $y$ is estimated as
\begin{equation}
    \label{eq__6}
    p(y|x_q) = 
    \frac{exp(-d(s\odot{\hat{a}_{x_q}},s\odot{\hat{c}_y}))}
    {\sum_{y' \in {\mathcal{Y}_s}}exp(-d(s\odot{\hat{a}_{x_q}},s\odot{\hat{c}_{y'}}))}
\end{equation}
where $\odot{}$ denotes the element-wise product operation. When $s^i$ is 1, the $i^{th}$ attribute is active and the distance along the corresponding dimension will affect the class membership probabilities calculated in the episode. Similarly when the state is 0, the attribute will not influence the classification. Therefore, only the relevant attributes selected in each episode are used in the classification.

The loss term for few-shot classification to train $g_{h}$ is formulated as:
\begin{equation}
    \label{eq__7}
    l_{cls} = - \sum_{x_q \in \mathcal{Q}_b}  \log p(y_{x_q}|x_q)
\end{equation}
The probabilities used in $l_{cls}$ are computed using Equation \ref{eq__6}.
Although the attribute selection mechanism can improve interpretability by decreasing the number of attributes participating in the few-shot recognition, this reduction is not guaranteed. Therefore, we also use $l_1$ norm of the binary states vector $s$ to further control and reduce the number of selected attributes in each episode. The final objective function to train $g_{h}$ is:
\begin{equation}
    \label{eq__8}
    l_{sel} = l_{cls} + \eta l_1
\end{equation}
where the hyperparameter $\eta$ specifies the weights of the loss term $l_1$. 

\subsection{Unknown Attribute Participation}
\label{unk_atts}

Human-friendly attributes may not always be able to properly perform few-shot learning due to their insufficiency and lack of quality. To address such situations and close the performance gap, we propose a method that allows trading interpretability for accuracy. We aim to improve accuracy by introducing \textit{unknown} attributes $\bar{a}$ to be used additional to the human-friendly attributes for few-shot classification. The new attributes are called unknown since they are not human-friendly; we learn them during training. 

In each episode, we attempt to balance the interpretability and accuracy by automatically deciding whether the unknown attributes should be involved or not. This is achieved by training a network $g_{u}$ that receives the human-friendly and unknown attributes and makes the above decision. $g_{u}$ aims to detect when the human-friendly attributes are not adequate and adding unknown attributes would be most effective. 

To learn unknown attributes suitable for few-shot classification, we train a network $f_{u}$ with the same structure as $f_{h}$ using the objective function $l_{cls}$ shown in Equation \ref{eq__7} . The probability in Equation \ref{eq__7} is calculated by Equation \ref{eq__4}. The trained network $f_{u}$ is then used to extract embedding features acting as unknown attributes for each sample. 

Although training $f_{u}$ using $l_{cls}$ can result in obtaining unknown attributes suitable for few-shot classification, they may have some overlap with human-friendly attributes. This overlap hurts the possibility of human test-time intervention on human-friendly attributes. Therefore, we also consider minimizing the mutual information \textit{I} between the learned human-friendly attributes $\hat{a}$ and unknown attributes $\bar{a}$ during training $f_{u}$:
\begin{equation}
    \label{eq__9}
    \begin{aligned}
        \min_{f_{u}} I(\bar{a}, \hat{a})
    \end{aligned}
\end{equation}

However, due to unavailability of a direct general-purpose procedure for optimizing the above problem in a high dimensional space, we leverage the mutual information neural estimator presented in \cite{pmlr-v80-belghazi18a} and consider an auxiliary network $f_I: \mathbb{R}^{A+\bar{A}}\rightarrow\mathbb{R}$  to estimate the mutual information $I(\bar{a}, \hat{a})$ by solving the following maximization problem:

\begin{equation}
    \label{eq__neural_mutual}
    \begin{aligned}
        I(\bar{a}, \hat{a}) \approx \max_{f_{I}} \mathbb{E} [f_I(\bar{a}, \hat{a})] - log(\mathbb{E} [exp(f_I(\bar{a}, \hat{a}))])
    \end{aligned}
\end{equation}

Learning the parameters of $f_I$ by maximizing Equation \ref{eq__neural_mutual} leads to acquiring an estimation of $I(\bar{a}, \hat{a})$. By substituting Equation \ref{eq__neural_mutual} into Equation \ref{eq__9}, we obtain the following minimax problem:

\begin{equation}
    \label{eq__10}
    \begin{aligned}
        l_{MI} = \min_{f_{u}} \max_{f_{I}} \mathbb{E} [f_I(\bar{a}, \hat{a})] - log(\mathbb{E} [exp(f_I(\bar{a}, \hat{a}))])
    \end{aligned}
\end{equation}
The final objective function to train $f_{u}$ is:
\begin{equation}
    \label{eq__10}
    \begin{aligned}
        l_{fake} = l_{cls} + \lambda l_{MI}
    \end{aligned}
\end{equation}
where $\lambda$ controls the mutual information minimization term. The total number of unknown attributes is the same as the total number of human-friendly attributes. The networks $f_{u}$ and $f_{I}$ are trained simultaneously. More details are provided in \ref{app_unk}.

\begin{table*}
\caption{Results on CUB and aPY with 5-way setting. Mean accuracy with 95\% confidence interval over 600 episodes sampled from novel classes is reported.}
\label{tab_1}
\centering
\begin{tabular}{lcccccc}
\hline
\multirow{2}{*}{\textbf{Method}} & \multicolumn{3}{c}{\textbf{CUB}}  & \multicolumn{3}{c}{\textbf{aPY}} \\
                     & \textbf{1-shot} & \textbf{3-shot} & \textbf{5-shot} & \textbf{1-shot} & \textbf{3-shot} & \textbf{5-shot}\\
\hline
\textbf{MatchingNet} & 60.3 $\pm$ 1 & 71.0 $\pm$ 0.8 & 74.2 $\pm$ 0.6 & 40.6 $\pm$ 0.7 & 53.3 $\pm$ 0.7 & 56.1 $\pm$ 0.7\\
\textbf{MAML} & 55.6 $\pm$ 1 & 63.4 $\pm$ 0.9 & 66.4 $\pm$ 0.8 &  40.7 $\pm$ 0.7 & 48.2 $\pm$ 0.7 & 49.4 $\pm$ 0.6\\
\textbf{RelationNet} & 60.3 $\pm$ 0.9 & 69.0 $\pm$ 0.7 & 72.1 $\pm$ 0.7 &  38.1 $\pm$ 0.7 & 46.7 $\pm$ 0.6 & 52.3 $\pm$ 0.7\\
\textbf{ProtoNet} & 53.8 $\pm$ 0.9 & 67.1 $\pm$ 0.7 & 71.4 $\pm$ 0.8 &  41.9 $\pm$ 0.8 & 51.0 $\pm$ 0.7 & 56 $\pm$ 0.7\\

\hline

\textbf{Ours} & 56.3 $\pm$ 0.9 & 66.7 $\pm$ 0.8 & 71.4 $\pm$ 0.8 & 40.6 $\pm$ 0.7 & 46.9 $\pm$ 0.7 & 53.5 $\pm$ 0.7\\

\end{tabular}
\end{table*}

\begin{table*}
\caption{Results on SUN and AWA with 5-way setting. Mean accuracy with 95\% confidence interval over 600 episodes sampled from novel classes are reported.}
\label{tab_2}
\centering
\begin{tabular}{lcccccc}
\hline
\multirow{2}{*}{\textbf{Method}} & \multicolumn{3}{c}{\textbf{SUN}}  & \multicolumn{3}{c}{\textbf{AWA}} \\
                     & \textbf{1-shot} & \textbf{3-shot} & \textbf{5-shot} & \textbf{1-shot} & \textbf{3-shot} & \textbf{5-shot}\\
\hline
\textbf{MatchingNet} & 63.2 $\pm$ 1 & 74.1 $\pm$ 0.8 & 78.3 $\pm$ 0.8 & 45.1 $\pm$ 0.9 & 54.4 $\pm$ 0.7 & 59.9 $\pm$ 0.7\\
\textbf{MAML} & 60.6 $\pm$ 1 & 68.6 $\pm$ 0.9 & 70.3 $\pm$ 0.9 & 35.1 $\pm$ 0.8 & 44.1 $\pm$ 0.8 & 46.9 $\pm$ 0.7\\
\textbf{RelationNet} & 62.9 $\pm$ 0.9 & 72.7 $\pm$ 0.8 & 77.7 $\pm$ 0.8 & 44.6 $\pm$ 0.8 & 52.9 $\pm$ 0.7 & 56.1 $\pm$ 0.7\\
\textbf{ProtoNet} & 63.7 $\pm$ 0.9 & 75.6 $\pm$ 0.7 & 78.6 $\pm$ 0.7 & 44.5 $\pm$ 0.8 & 53.3 $\pm$ 0.8 & 59.1 $\pm$ 0.7\\

\hline

\textbf{Ours} & 61.4 $\pm$ 1 & 72.2 $\pm$ 0.9 & 76.5 $\pm$ 0.7 & 42.6 $\pm$ 0.7 & 51.5 $\pm$ 0.8 & 56.1 $\pm$ 0.7\\

\end{tabular}
\end{table*}






The structure of $g_u$ is similar to the attribute selector network $g_{h}$ with the difference that this network will decide on the participation of the unknown attributes in each episode. Therefore, its output will be a scalar value. The final loss function for the proposed model with $g_{u}$ is similar to Equation \ref{eq__8} where $l_{1}$ is the output of $g_{u}$ and $\beta$ is the corresponding hyper-parameter. Moreover, $l_{cls}$ is calculated in the mixed space of unknown attributes and human-friendly attributes. The human-friendly attributes and unknown attributes are concatenated to construct this mixed space. For each sample, the output value of $g_{u}$, which is a scalar, is used as the weight for each of the unknown attributes and the binary weight vector returned by $g_{h}$ is used to weight the human-friendly attributes.  
Detailed explanation on training $g_{u}$ is also provided in \ref{app_unk}. It is noteworthy that the recent few-shot learning methods including MetaBaseline \cite{chen2021meta} and NegMargin \cite{liu2020negative} can be adapted to achieve unknown attributes. We leave this for future work.

\section{Experiments and Evaluation}
\label{eval}

In this section, we first describe the datasets, experimental settings and implementation details. Then we present the results of the proposed interpretable method and compare it with a number of previously state-of-the-art black-box FSL methods.

\subsection{Experimental Setup}

\textbf{Datasets and experimental settings:} We conduct experiments on four popular computer vision datasets with available attributes including Caltech-UCSD Birds-200-2011 (CUB) \cite{WahCUB_200_2011}, Animals with Attributes (AWA) \cite{Xian2018}, SUN Attributes (SUN) \cite{Patterson2012} and  Attribute Pascal and Yahoo (aPY) \cite{5206772}. The statistics of these datasets and their setup are provided in \ref{app_data}.

The experiments are performed on three widely used 5-way 1-shot, 5-way 3-shot and 5-way 5-shot settings. For each class, a query set consists of 16 samples is constructed in each episode for CUB, AWA and aPY. For SUN, the query set contains 10 samples due the lower number of instances in each class. 
In each experiment, we report the mean accuracy with 95\% confidence interval over 600 episodes sampled from novel classes which is widely used as performance measure of FSL approaches.

\textbf{Implementation Details:}
As the attribute predictor network, we use a four blocks convolutional neural network similar to Conv-4 \cite{Proto} which is one of the most adopted backbone networks for FSL. In the original network, each block consists of a 3 × 3 convolution with 64 filters, followed by a batch normalization layer, a Relu non-linearity and a 2 × 2 max pooling layer. However, we change the number of filters to achieve feature maps with the number of channels equal to the total number of attributes in the final block. The Conv-4 is followed by a global average pooling. Then the features are fed to a linear layer with tanh activation function for its more symmetric properties and a wider output range compared to sigmoid activation function. The output is then normalized to have a range of [0,1], compatible with binary cross entropy.
The attribute selector network consists of a one layer Bi-LSTM with 100 features in the hidden state, followed by a one-layer perceptron with sigmoid activation function. The Gumbel softmax sampler then uses the output of the selector network. The value of $\eta$ is specified in each subsection. Additionally, we use Euclidean distance as the distance function in all experiments.
More details regarding our implementation are available in \ref{app_imp}.

\subsection{Performance comparison}

In this section, we evaluate the performance of our method from two different angles: 1) Accuracy 2) Decision alignment with human understanding. The results are compared with previously state-of-the-art FSL methods, including MatchingNet \cite{vinyals2016matching}, MAML \cite{finn2017model}, RelationNet \cite{sung2018learning} and ProtoNet \cite{Proto}. Arguably, these are seminal works in the area of FSL.

\textbf{Accuracy:} As discussed previously, attributes act as bottleneck in our method to ensure decisions can be traced back to human-friendly attributes. We seek to maximize the accuracy but within the interpretability constraint mentioned above. For the above reason, the proposed method is not expected to always achieve a higher accuracy than the black-box methods that do not have any explainability constraints. Despite this, the accuracy of the proposed method is competitive with that of the black-box models. The results are presented in Tables \ref{tab_1} and \ref{tab_2}. To have a fair comparison, all methods are implemented using the same backbone network. Note that we did not use unknown attributes in this experiment. Furthermore, the L1 norm is not applied to the output of $g_{h}$ (i.e. $\eta = 0$).


 

\begin{figure}
  \centering 
  \includegraphics[width=0.95\linewidth]{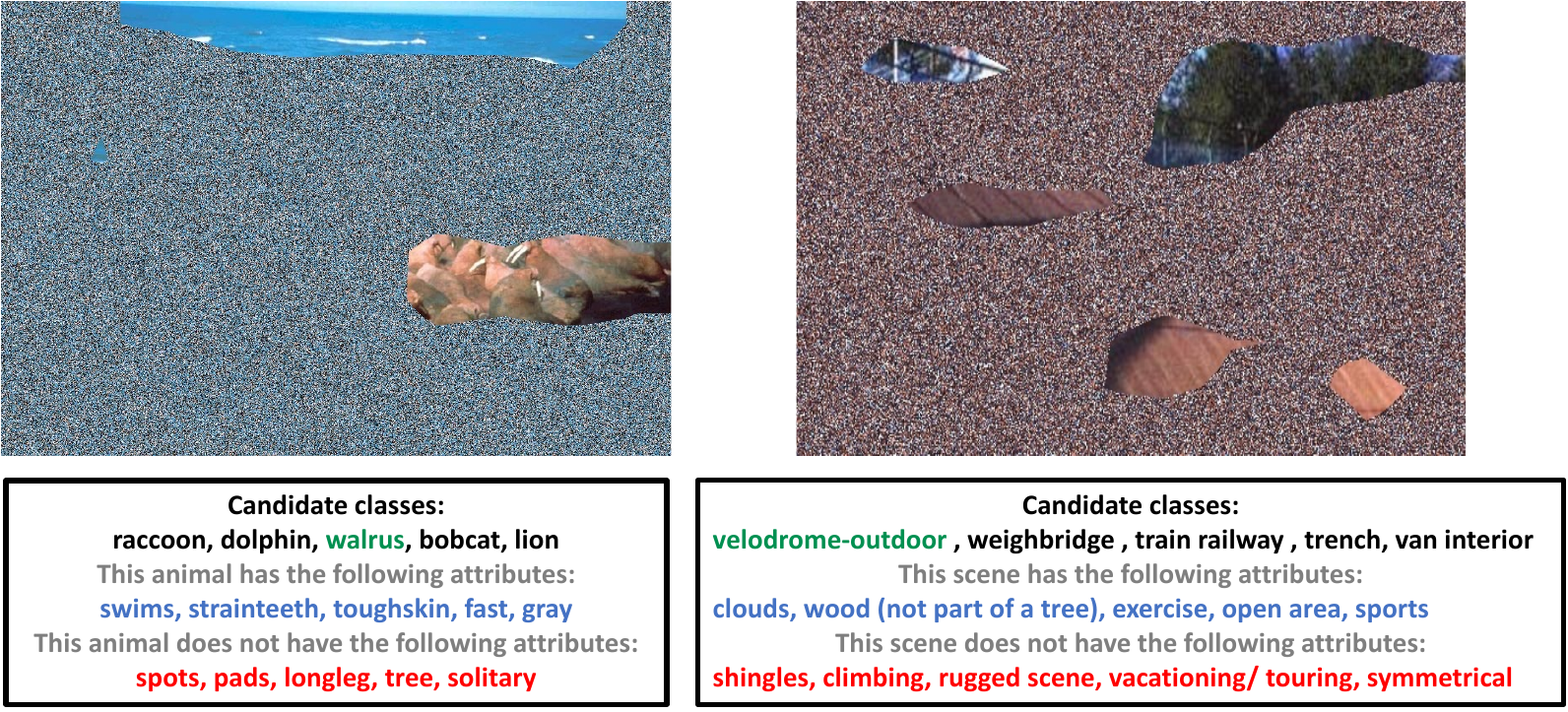}
  \caption{Examples of partially visible query images together with the selected attributes. In the absence of the attributes, it is hard for human to tell the image class only based on the masked image. Providing attributes makes it easier for human to predict the image class.}
  \label{fig.examples}
\end{figure}

\textbf{Decision Alignment with Human Understanding:} While various techniques exist for explaining the decision of deep learning-based models, there is no standardized approach for quantitatively comparing the explainability of different models and how well they align with human understanding. Therefore, we designed and implemented an experiment to evaluate the consistency of the models decision with human interpretation. For each query sample, we extract its salient regions using Grad-CAM \cite{selvaraju2017grad} and ask non-expert individuals to use that information to classify images. Since our model can predict human-friendly attributes, we also provide this information for our own model. By evaluating the performance of the individuals in predicting the actual class using the provided model-specific information, we can measure and compare the degree of explainability of the models.


For this experiment, we first sample 200 episodes from novel classes in each setting and randomly select one query image from the query set of each of these 200 episodes, resulting in a total of 200 samples. For each model, we obtain the grad-cam activation maps of the query images with respect to the final decision. Then, we use the values of the grad-cam activation maps to create a partially visible version of the query images with only salient regions being visible. Figure \ref{fig.examples} shows examples of such masked images. Since pixels with higher grad-cam activation value indicate the locations that are more important to the model in deciding the class of a query image, we only keep the pixels with a Grad-CAM activation value higher than a threshold and shuffle the remaining pixels in the image. This threshold is set to $0.5$ for all models and settings. Each masked query image contains the regions of the original image that were most important for the final decision of the model. Therefore, it is a good indicator of the decision explainability of the corresponding model.

We used Amazon Mechanical Turk (AMT) to recruit non-expert individuals (MTurk workers) for this experiment. Each experiment includes 200 tasks for each model. Each task represents one few-shot episode in which we first present workers with the support set images to familiarize them with the classes participating in the current episode, then we show the masked query image and ask them to choose the most appropriate class.

\begin{table}
\caption{Accuracy of the 5-way 1-shot classification carried out by the workers after being given the masked images, or in the case of the proposed method, the masked images together with the relevant attributes. The degree that the salient regions are useful to human for predicting the class of the images is an indicator of the degree of alignment between the model and human understanding.}
\label{tab_exp1}
\centering
\begin{adjustbox}{width={0.48\textwidth},keepaspectratio}
\begin{tabular}{lcccc}
\hline
\multirow{2}{*}{\textbf{Method}} & \multicolumn{3}{c}{\textbf{Dataset}} & \multirow{2}{*}{\textbf{Average}}\\
                     & \textbf{APY} & \textbf{SUN} & \textbf{AWA} & \\
\hline
\textbf{MatchingNet} & 61.3 $\pm$ 1.9 & 75.2 $\pm$ 1.7 & 73.2 $\pm$ 1.9& 69.9\\
\textbf{MAML} & 60.3  $\pm$ 2.1 & 75.6 $\pm$ 2 & 77.8 $\pm$ 1.8& 71.2\\
\textbf{RelationNet} & 61.2 $\pm$ 1.8 & 77.2 $\pm$ 1.8 & 68.7 $\pm$ 2& 69\\
\textbf{ProtoNet} & 81.6 $\pm$ 1.3 & 79.5 $\pm$ 1.6 & 68.7 $\pm$ 2.1& 76.6\\

\hline

\textbf{Ours \footnotesize{(without attributes)}} & 76.9 $\pm$ 1.9 & 81.3 $\pm$ 1.6 & 82.4 $\pm$ 1.4& 80.2\\

\textbf{Ours} & 82 $\pm$ 1.6 & 82.3 $\pm$ 1.6 & 83.2  $\pm$ 1.5& 82.5\\

\end{tabular}
\end{adjustbox}
\end{table}

\begin{table}
\caption{Accuracy of the 5-way 3-shot classification carried out by the workers after being given the masked images, or in the case of the proposed method, the masked images together with the relevant attributes. The degree that the salient regions are useful to human for predicting the class of the images is an indicator of the degree of alignment between the model and human understanding.}
\label{tab_exp2}
\centering
\begin{adjustbox}{width={0.48\textwidth},keepaspectratio}
\begin{tabular}{lcccc}
\hline
\multirow{2}{*}{\textbf{Method}} & \multicolumn{3}{c}{\textbf{Dataset}} & \multirow{2}{*}{\textbf{Average}}\\
                     & \textbf{APY} & \textbf{SUN} & \textbf{AWA} &\\
\hline
\textbf{MatchingNet} & 63.1 $\pm$ 1.9 & 76.5 $\pm$ 1.9 & 67.6 $\pm$ 2.1 & 69.1\\
\textbf{MAML} & 59.7 $\pm$ 2.1 & 71.5 $\pm$ 2 & 61.3 $\pm$ 2& 64.2\\
\textbf{RelationNet} & 68.6 $\pm$ 1.8 & 76.8 $\pm$ 1.7 & 63.7 $\pm$ 1.8& 69.7\\
\textbf{ProtoNet} & 82.4 $\pm$ 1.2& 77.2 $\pm$ 1.6 & 72.6 $\pm$ 1.7 & 77.4\\

\hline

\textbf{Ours \footnotesize{(without attributes)}} & 77 $\pm$ 1.8& 80.9 $\pm$ 1.8 & 82.3 $\pm$ 1.5 & 80.1\\

\textbf{Ours} & 82.6 $\pm$ 1.6& 81.6 $\pm$ 1.8 & 83.9 $\pm$ 1.6 & 82.7\\

\end{tabular}
\end{adjustbox}
\end{table}

\begin{table}
\caption{Accuracy of the 5-way 5-shot classification carried out by the workers after being given the masked images, or in the case of the proposed method, the masked images together with the relevant attributes. The degree that the salient regions are useful to human for predicting the class of the images is an indicator of the degree of alignment between the model and human understanding. } 
\label{tab_exp3}
\centering
\begin{adjustbox}{width={0.48\textwidth},keepaspectratio}
\begin{tabular}{lcccc}
\hline
\multirow{2}{*}{\textbf{Method}} & \multicolumn{3}{c}{\textbf{Dataset}} & \multirow{2}{*}{\textbf{Average}}\\
                     & \textbf{APY} & \textbf{SUN} & \textbf{AWA} &\\
\hline
\textbf{MatchingNet} & 70.8 $\pm$ 1.9 & 76.1 $\pm$ 1.8 & 73.9 $\pm$ 1.9 & 73.6\\
\textbf{MAML} & 64.2 $\pm$ 1.9 & 67.1 $\pm$ 2 & 67.2 $\pm$ 2.1& 66.2\\
\textbf{RelationNet} & 78.7 $\pm$ 1.5 & 72.6 $\pm$ 1.8 & 67.3 $\pm$ 2& 72.9\\
\textbf{ProtoNet} & 82.4 $\pm$ 1.5 & 83 $\pm$ 1.6 & 77.5 $\pm$ 1.8& 81\\

\hline

\textbf{Ours \footnotesize{(without attributes)}} & 78.7 $\pm$ 1.5 & 82.9 $\pm$ 1.6& 79.2 $\pm$ 1.7& 80.3\\

\textbf{Ours} & 83.7 $\pm$ 1.7 & 84.3 $\pm$ 1.5& 82.2 $\pm$ 1.6& 83.4\\

\end{tabular}
\end{adjustbox}
\end{table}

For our model, in addition to the masked image, we provide the workers with the attributes predicted by our attribute predictor network. Due to the large number of attributes, we first utilize the attribute selector network to filter out irrelevant attributes. Next, we assign a ranking to the remaining attributes according to how much of an impact they have on the distance between the query and the prototype of the selected class. The distance with and without the predicted value of each attribute is computed, and the difference between the two is utilized as a measure of importance. Finally, we pick the five most important present attributes and five most important absent attributes to form the final list of the attributes that will be presented to workers. Figure \ref{fig.examples} shows two examples of masked query images together with the attributes selected by our method. Each task was assigned to five workers and the classification accuracy of the workers on all the tasks corresponding to each model is averaged and reported as the indicator of the decision explainability of that model. 

To ensure the quality of the collected results, we implemented a two-step quality control protocol. In the first step, which occurs prior to taking the primary tasks, we ask workers to take a set of manually created dataset-specific qualification tasks and then allow them to take the primary tasks only if they succeed to pass the qualification tasks with 100\% accuracy. This collection of qualification tasks includes tasks similar to the primary ones with the difference that the class is certainly recognizable by observing the altered query image and the presented attributes. As the second step of the quality control protocol, we incorporate a set of manually created validation tasks alongside the main tasks which are easier to take and serve validating performance of each worker. The primary tasks performed by each worker are approved only if they were able to achieve 100\% accuracy on the anonymous validation tasks.
 
 Since the fine-grained bird classification in the CUB dataset requires prior expertise and proficiency in various bird species, it is not suitable for this experiment which is based on AMT workers. Therefore, we excluded it from this experiment and performed the experiment only on the other three datsets. The results are shown in Tables \ref{tab_exp1}, \ref{tab_exp2} and \ref{tab_exp3}. To further examine the usefulness of the attributes to human, we also report the results of the proposed method without attributes. It can be seen that the proposed method has achieved the highest task accuracy among all competitors in all settings when both images and attributes are provided to workers. This shows that the decision of our model is more aligned with human understanding compared to the other methods.

 In 5-way 1-shot setting, the proposed method achieves the best performance with an average accuracy of 82.5\% over the three datasets. This is followed by our approach without attributes and ProtoNet, with average accuracies of 80.2\% and 76.6\%, respectively. The same pattern is observed in 5-way 3-shot setting, where our approach continues to lead with an average accuracy of 82.7\%. The proposed method without attributes and ProtoNet achieve average accuracies of 80.1\% and 77.4\%, respectively. In 5-way 5-shot setting, our model outperforms all competitors, obtaining an average accuracy of 83.4\%. ProtoNet is the second-best performer, followed by our model without attributes with a 0.7\% difference. The lowest average accuracy in both 5-way 3-shot and 5-way 5-shot settings belongs to MAML, while RelationNet ranks last in 5-way 1-shot setting.

 It can be seen that when the attribute information is not shown to workers, their ability to correctly predict the true class significantly degrades. This shows that providing predicted attributes, in addition to the masked images, helps human makes more accurate predictions. Moreover, by comparing the results of decision alignment with human understanding and the accuracy of the models in previous subsections, we can infer that a higher accuracy does not necessarily lead to better decision explainability in the human view.

In the next section, we evaluate the effectiveness of the proposed attribute selector. We provide the results for the attribute predictor network in \ref{app_pre}.

\subsection{Attribute selector evaluation}

The attribute selector network aims to select a subset of the most relevant attributes in each few-shot episode. In this section, we examine the effectiveness of the attribute selector network  
by comparing it with a base model that does not do attribute selection. Similar to the previous section, we do not leverage unknown attributes in this experiment. The base model consists of just $f_{h}$ and performs few-shot learning using Equation \ref{eq__4}. Therefore, all attributes participate in few-shot classification. The results of these experiments are presented in the first column of Tables \ref{tab_6}, \ref{tab_7}, \ref{tab_8}, \ref{tab_9}. 
We also examine the effect of the ($l_1$) loss by examining different values of $\eta$ selected from the values [$0$,$10^{-5}$, $10^{-4}$, $10^{-3}$]. $\eta=0$ correspond to the proposed model without the ($l_1$) loss. We provided the average number of selected attributes in addition to the mean accuracy. Compared to the baseline model, the proposed method has reduced the number of attributes by 48, 38.4, 27.4 and 42\% while improving the accuracy by an average of 2.8, 1.8, 1.2 and 0.4\% on CUB, aPY, SUN and AWA datasets, respectively. 
Increasing $\eta$ significantly reduces the average number of selected attributes at the cost of only a slight drop in accuracy. For example, increasing $\eta$ from 0 to $10^{-3}$ reduces the average number of selected attributes by 62, 23, 33, and 33\% on CUB, aPY, SUN and AWA datasets, respectively, while hindering the accuracy by only 1.17, 1.14 , 0.5\%, and in the case of AWA the accuracy actually improved by 0.66\%. Furthermore, the number of selected attributes decreases when the size of the support set is reduced from K=5 to 1. This is an interesting behavior that demonstrates how the proposed method attempts to avoid the overfitting problem by choosing fewer attributes. We provide qualitative examples regarding attribute selection mechanism in \ref{app_selected_attributes}.

\begin{table}
\caption{Evaluating the impact of attribute selector network on CUB in terms of average accuracy and number of attributes. The base model does not have $g_{h}$. The total number of attributes is 312 for this dataset.} 
\label{tab_6}
\centering
\begin{tabular}{cc|ccccc}
\hline
\multicolumn{2}{c|}{\multirow{2}{*}{\textbf{Setting}}} & \multirow{2}{*}{Base} & \multicolumn{4}{c}{\textbf{$\eta$}} \\           
  &   &  & \textbf{$0$} & \textbf{$10^{-5}$} & \textbf{$10^{-4}$} & \textbf{$10^{-3}$}\\
\hline
\multirow{2}{*}{\textbf{1-shot}} & \textbf{Acc} & 52.4 & 56.3 & 56.6 & 55.9 & 55\\
   & \textbf{\# Att} & 312 & 97.1 & 93.5 & 80.6 & 44.1\\

\hline
\multirow{2}{*}{\textbf{3-shot}} & \textbf{Acc} & 64.1 & 66.7 & 66.8 & 66.3 & 65.7\\
    & \textbf{\# Att} & 312 & 170.6 & 154.7 & 133.4 & 67.1\\

\hline

\multirow{2}{*}{\textbf{5-shot}} & \textbf{Acc} & 69.5 & 71.4 & 71.4 & 73.5 & 70.2\\

    & \textbf{\# Att} & 312 & 219.1 & 186.5 & 148.9 & 73\\

\end{tabular}
\end{table}

\begin{table}
\caption{Evaluating the impact of attribute selector network on aPY in terms of average accuracy and number of attributes. The base model does not have $g_{h}$. The total number of attributes is 64 for this dataset.}
\label{tab_7}
\centering
\begin{tabular}{cc|ccccc}
\hline
\multicolumn{2}{c|}{\multirow{2}{*}{\textbf{Setting}}} & \multirow{2}{*}{Base} & \multicolumn{4}{c}{\textbf{$\eta$}} \\          
  &   &  &\textbf{$0$} & \textbf{$10^{-5}$} & \textbf{$10^{-4}$} & \textbf{$10^{-3}$}\\
\hline
\multirow{2}{*}{\textbf{1-shot}} & \textbf{Acc} & 40.0 & 40.6 & 40.8 & 40.8 & 39.9\\

   & \textbf{\# Att} & 64 & 36.4 & 34.2 & 30 & 25.8\\

\hline
\multirow{2}{*}{\textbf{3-shot}} & \textbf{Acc} & 45.2 & 46.9 & 46.4 & 46.3 & 46\\
    & \textbf{\# Att} & 64 & 38.5 & 37.5 & 34.1 & 32.6\\
\hline
\multirow{2}{*}{\textbf{5-shot}} & \textbf{Acc} & 50.3 & 53.5 & 52 & 52 & 51.7\\
    & \textbf{\# Att} & 64 & 43.4 & 39.1 & 35.9 & 33\\
\end{tabular}
\end{table}

\begin{table}
\caption{Evaluating the impact of attribute selector network on SUN in terms of average accuracy and number of attributes. The base model does not have $g_{h}$. The total number of attributes is 102 for this dataset.}
\label{tab_8}
\centering
\begin{tabular}{cc|ccccc}
\hline
\multicolumn{2}{c|}{\multirow{2}{*}{\textbf{Setting}}} & \multirow{2}{*}{Base} & \multicolumn{4}{c}{\textbf{$\eta$}} \\             
  &   &  & \textbf{$0$} & \textbf{$10^{-5}$} & \textbf{$10^{-4}$} & \textbf{$10^{-3}$}\\
\hline
\multirow{2}{*}{\textbf{1-shot}} & \textbf{Acc} & 60.4 & 61.4 & 59.6 & 60.3 & 60.3\\
   & \textbf{\# Att} & 102 & 53.48 & 50.7 & 50.4 & 38.4\\
\hline

\multirow{2}{*}{\textbf{3-shot}} & \textbf{Acc} & 72.0 & 72.2 & 72.8 & 72.9 & 72.3\\
    & \textbf{\# Att} & 102 & 82.4 & 80.7 & 77 & 51.9\\
\hline

\multirow{2}{*}{\textbf{5-shot}} & \textbf{Acc} & 74.1 & 76.5 & 76 & 75.4 & 76\\
    & \textbf{\# Att} & 102 & 86.2 & 83.9 & 81.2 & 58\\

\end{tabular}
\end{table}

\begin{table}
\caption{Evaluating the impact of attribute selector network on AWA in terms of average accuracy and number of attributes. The base model does not have $g_{h}$. The total number of attributes is 85 for this dataset.}
\label{tab_9}
\centering
\begin{tabular}{cc|ccccc}
\hline
\multicolumn{2}{c|}{\multirow{2}{*}{\textbf{Setting}}} & \multirow{2}{*}{Base} & \multicolumn{4}{c}{\textbf{$\eta$}} \\          
  &   &  & \textbf{$0$} & \textbf{$10^{-5}$} & \textbf{$10^{-4}$} & \textbf{$10^{-3}$}\\
\hline
\multirow{2}{*}{\textbf{1-shot}} & \textbf{Acc} & 41.8 & 42.6 & 42.6 & 43.5 & 42.7\\
   & \textbf{\# Att} & 85 & 39.8 & 38.7 & 38.2 & 29.5\\
\hline
\multirow{2}{*}{\textbf{3-shot}} & \textbf{Acc} & 51.8 & 51.5 & 52.3 & 51.2 & 53.3\\
    & \textbf{\# Att} & 85 & 52.3 & 51.9 & 46.7 & 35.3\\
\hline

\multirow{2}{*}{\textbf{5-shot}} & \textbf{Acc} & 55.3 & 56.1 & 55.9 & 56 & 56.2\\

    & \textbf{\# Att} & 85 & 55.7 & 50.7 & 48.1 & 34\\

\end{tabular}
\end{table}

\begin{table*}
\caption{Evaluating the effectiveness of the proposed $g_{u}$ on AWA and CUB datasets in 5-way 1-shot. Mean accuracy with 95\% confidence interval over 600 episodes sampled from novel classes is reported.}
\label{tab_10}

\centering
\begin{adjustbox}{width={\textwidth},keepaspectratio}
\begin{tabular}{l|ccc|ccc|ccc|ccc}

Dataset &  \multicolumn{3}{c}{\textbf{CUB10\%}} & \multicolumn{3}{c}{\textbf{CUB}} &  
\multicolumn{3}{c}{\textbf{AWA10\%}} &
\multicolumn{3}{c}{\textbf{AWA}} \\
\hline
 $\beta$  &  1.75  & 1.7  & - & 0.7 & 0.8 & - & 0.3 & 0.4 & - &  0.03  & 0.04 & -\\
 \% of human-friendly episodes & 28.3 & 74.8 & 100 & 50.3 & 84 & 100 & 35.5 & 69.3 & 100 & 48 & 89.3 & 100\\

Accuracy with $g_{u}$  & 58.6 $\pm$ 0.9 & 52 $\pm$ 1 & 48.1 $\pm$ 1 & 60.5 $\pm$ 0.9 & 57.5 $\pm$ 0.9 & 56.3 $\pm$ 0.9 & 40.8 $\pm$ 0.8 & 38.3 $\pm$ 0.8 & 36.3 $\pm$ 0.8 & 45.2 $\pm$ 0.8 & 43.1 $\pm$ 0.7 & 42.6 $\pm$ 0.7\\

\end{tabular}
\end{adjustbox}
\end{table*}


\subsection{Closing the Accuracy Gap}
Looking at Table 1 and 2, it can be seen that among the four datasets, the biggest performance gap between the proposed method and the bests black-box method is on CUB and AWA. This could be because the attributes or their quality are not adequate to distinguish the classes. In this section, we provide the results of incorporating unknown attributes as disused in Section \ref{unk_atts}. 

The results of this experiment on CUB and AWA in 5-way 1-shot setting are shown in Table \ref{tab_10}. To further evaluate how well the proposed approach can compensate for accuracy when the attributes are incomplete, we also report the results on two new versions of CUB and AWA with only 10\% of the attributes randomly selected, namely CUB10\% and AWA10\%, respectively. $\beta$ indirectly controls the proportion of human-friendly episodes among all 600 sampled episodes of novel classes. We report the results of the models trained with two different values of $\beta$ on each dataset. We consider $\{1.7,1.75\}$, $\{0.7,0.8\}$, $\{0.3,0.4\}$ and $\{0.03,0.04\}$ for CUB10\%, CUB, AWA10\% and AWA, respectively. It is noteworthy that these numbers are selected based on the validation set to give us a mid-high and mid-low percentage of human-friendly episodes.

On the AWA dataset, by using $g_{u}$ we were able to get closer to the accuracy of the best black-box model (45.1\%) by improving the accuracy from 42.6 to 43.1, and yet making interpretable decisions in 89.3\% of the episodes. For a smaller value of $\beta$, the proposed approach surpassed the best model by achieving an accuracy of 45.2\% while 48\% episodes still were handled by only human-friendly attributes. On the CUB dataset, we were able to improve the accuracy from 56.3 to 60.5 and consequently not only close the gap but also outperform the best black-box model while making interpretable decisions in more than half of the episodes. Finally, it can be seen that the accuracy promotion by utilizing unknown attributes is more significant when the number of human-friendly attributes is decreased to 10\% in AWA10\% and CUB10\% datasets.

\subsection{Preferability of human-friendly attributes over unknown attributes}
As seen in the previous subsection, the performance of the framework can be improved by utilizing unknown attributes when the known attributes are incomplete. Even in the episodes that the model leverages unknown attributes, the entire package is still interpretable since unknown attributes can be explained using saliency methods such as Grad-CAM \cite{selvaraju2017grad}. Examples of Grad-Cam visualization for different unknown attributes of CUB and AWA datasets are provided in \ref{app_unk}. Interestingly, the unknown attributes desire visually fine-grained attributes including body parts that can very well complement abstract known attributes.

\begin{figure}
  \centering 
  \includegraphics[width=\linewidth]{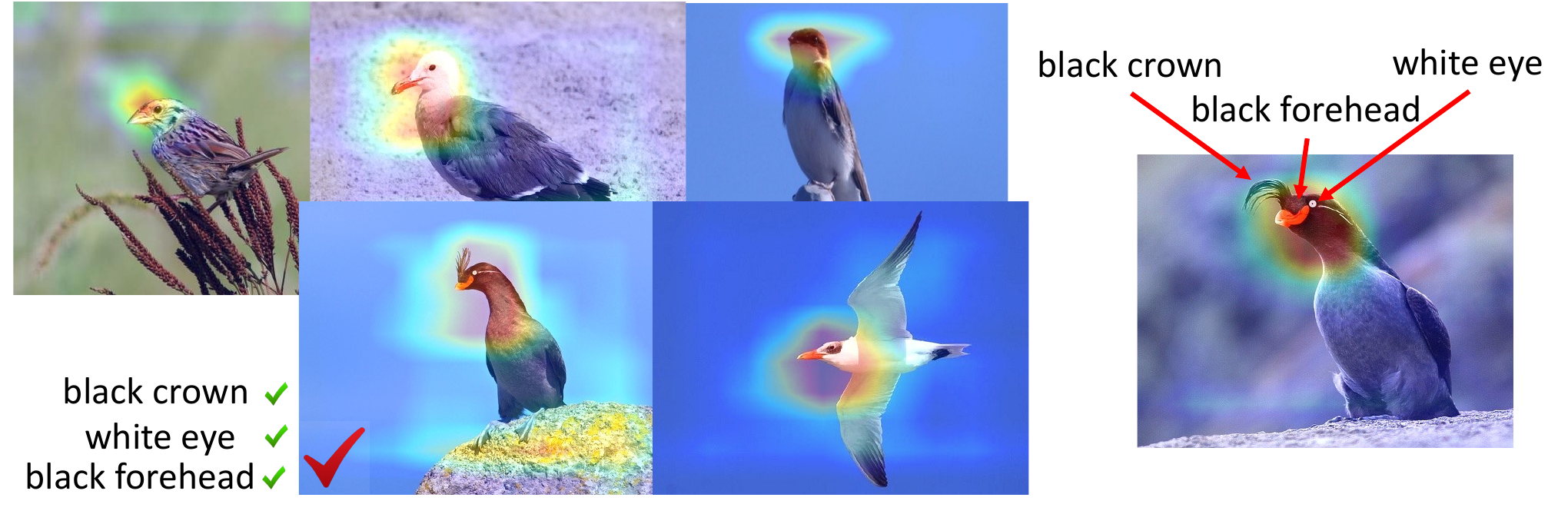}
  \caption{The Grad-cam visualization of an unknown attribute for a support set (left) and a query example (right) on CUB dataset. While heatmaps provide spacial localisation, they fail to indicate what exactly the model has focused on in a region. For example, is it beak shape, beak color, eye color? The human-friendly attributes detected by the proposed method can very well address this shortcoming.}
  \label{fig.grad2}
\end{figure}

Although the utilized saliency explanation can reveal useful information about what the model has focused on for the final decision, there is no guarantee regarding the reliability of the explanation \cite{adebayo2018sanity} and sometimes they can be misleading. This is a limitation of such approaches; it is not a phenomenon specific to our unknown attribute learning paradigm. Therefore, all the black-box models that utilize similar post-hoc interpretability approaches are facing the same challenge. Even if we assume this kind of explanation is reliable, it is often difficult to interpret the explanation in the direction of the model's decision, especially for fine-grained classification tasks. For example, Figure \ref{fig.grad2} shows the Grad-CAM of an unknown attribute for a support set and a query sample on the CUB dataset. Although saliancy maps provide spacial localization, it is not apparent what exactly the model has focused on in that region; is it beak shape, beak color, eye color? On the other hand, the proposed method was able to successfully recognize the human-friendly attributes that are useful for classification of the query sample including white eye, black forehead, and black crown. This highlights the advantages of the proposed method over attempting to produce post-hoc explanations for a black-box model.

\subsection{Human Intervention}

In this subsection, we consider the capability of our framework in enabling interaction with human by performing intervention on misclassified query samples. In conventional concept bottleneck models (CBM), a human can interact with the model by identifying incorrect predicted concepts for miclassified instances and rectify them to see how the final decision will change. This type of intervention is appropriate for CBM since it employs a parametric model for classification on top of the concept bottleneck layer. However, since our model uses a non-parametric approach in its classification phase and its decision also relies on the predicted concepts of the model for other instances (support set), correcting the concept prediction just for the query sample can not reflect the actual effect of the concept in updating final decision. Therefore, we design and present a new form of intervention applicable to our FSL model. 

To intervene an arbitrary attribute of a misclassified query instance, the user will observe this sample and the prototypes of different classes in the current episode and decides on the prototype most similar to the query instance in terms of the selected attribute. Then, the query will be moved toward the prototype but only along the dimension corresponding to the attribute in decision space. Since the decision space is the same as the attribute space in our model, the user rectifies the predicted distance between query instance and the prototypes in terms of the attribute that is being intervened.

There is no real-world human interaction in this experience. Instead, we rely on ground-truth annotations to simulate intervention and assume that real-world human intervention would be consistent with the ground-truth annotations. Therefore, to rectify the distance between a misclassified query sample and the prototypes, the prototype with the ground-truth attribute value same as the query sample is identified and the query sample is moved toward this prototype by setting its attribute value equal to the predicted attribute value for the identified prototype in the decision space. If there is more than one prototype with the same ground-truth attribute value, the predictions for all of them are averaged and will be used as the attribute value of the query sample.

We do intervention on the framework performing in joint attributes space (with unknown attributes) and when only human-friendly attributes are used (without unknown attributes). 600 episodes are sampled from novel classes and average accuracy before and after intervention with two intervention ratio (5\% and 10\%) are reported. It should be noted that in this experiment, unknown attributes are employed in all 600 episodes when evaluating intervention in joint attributes space. The results of this experiment are presented in Table \ref{tab_11}. In summary, the intervention improved the performance in all settings, obtaining an average improvement of 2.8\% and 2.65\% on CUB and AWA, respectively. The promotion was more significant with a higher ratio of intervention ($10\%$) in all cases and in human-friendly attribute space compared to joint attribute space. The complete results are provided in \ref{app_intervention}.


\section{Conclusion}

In this paper, we proposed an inherently interpretable few-shot learning method based on human-friendly attributes. During inference, the proposed method does not need the attributes. Attributes are extracted from input. Moreover, an attribute selection mechanism was proposed to filter out irrelevant attributes in each few-shot learning episode which in turn improves the accuracy. The proposed interpretable method achieved results on par with the previously state-of-the-art black-box few-shot learning methods. We also proposed a method that aimed to close the accuracy gap with the black-box models by automatically detecting the episodes that can benefit the most from unknown learned attributes. We demonstrated that this approach is effective in closing the performance gap while delivering interpretable FSL classification in a substantial number of episodes. 

\section*{Acknowledgement}

This work was partially supported by the Natural Sciences and Engineering Research Council of Canada, and Digital Research Alliance of Canada.

\section*{CRediT authorship contribution statement}

\textbf{Mohammad Reza Zarei:} Conceptualization, Methodology, Software, Validation, Writing - original draft. \textbf{Majid Komeili:} Conceptualization, Methodology, Supervision, Writing - review \& editing.

\appendix

\section{Datasets and Settings Description}
\label{app_data}

CUB is a fine-grained dataset containing 11,768 images from 200 bird classes. For each image, 312 binary attributes are provided in this dataset. Following the protocol proposed by \cite{chen2018a}, we divide the dataset into the same 100 base classes, 50 validation and 50 novel classes. AWA consists of 37,322 animal images from 50 different categories, with 85-dimensional class-level attribute vectors. We use the first 30 categories as base classes, the next 10 classes as validation and the last 10 categories as novel classes. SUN is a fine-grained scene dataset with 14340 images from 717 different classes. Each image is provided with 102 attribute. For each attribute, a continuous value in range of [0,1] is calculated by averaging the votes of three annotators. We round each value to achieve binary attributes. Furthermore, we divide this dataset into 580 base classes, 65 validation and 72 novel classes. aPY is a course-grained dataset containing 15,339 images from 32 classes, 20 Pascal and 12 Yahoo classes. All Yahoo classes are used as novel ones and 15 Pascal classes are used as base classes. The remaining 5 Pascal classes are used as validation categories. Each image in this dataset is provided with 64 attributes. 

\section{Additional Implementation Details}
\label{app_imp}

We apply data augmentation including color jittering and random horizontal flipping. Furthermore, the input size of 84 × 84 is used for all datasets. Both attribute predictor network and attribute selector network are trained using Adam optimizer with a learning rate equal to $10^{-3}$.

The attribute predictor network $f_{h}$ is trained for 100 epochs with mini batch size of 128 on all datasets. After each epoch, the attribute prediction performance on the samples of validation classes is measured. Then, the model with the best performance on validation classes is selected as the final network. 

The attribute selector network $g_{h}$ is trained for 200,000 episodes, evaluated on 600 sampled episodes of validation classes every 500 episodes. The model achieving the best performance on the validation episodes will be the final network. The temperature parameter of Gumbel Softmax estimator ($\tau$) in $g_{h}$ is initialized to 4, halved every 12500 episodes during training until reaching to 0.5.

We report the results of our framework with various values of $\eta$ selected from [$0$,$10^{-5}$, $10^{-4}$, $10^{-3}$]. These numbers are picked to cover a wide range of reduction rate in the average percentage of participated attributes in episodes and selected based on the performance of the framework on validation classes.

\begin{algorithm*}
    \caption{Training $f_{u}$ to learn unknown attributes}
    \label{algo1}
\label{alg:1}

    \textbf{Input:} $f_{u}$ and $f_{I}$ with their initialized weights $\theta_{f_{u}}$ and $\theta_{f_{I}}$, trained $f_{h}$ \\
    \For{$E_1$ episodes}{
        Draw $N$ samples ($\mathcal{S}$) from $\mathcal{D}_b$\\
        Obtain human-friendly attributes for $\mathcal{S}$ using $f_{h}$: {$\hat{\mathcal{A}}_S = \{\hat{a}_x| x \in \mathcal{S}\} $}\\
        Obtain unknown attributes for $\mathcal{S}$ using $f_{u}$: {$\bar{\mathcal{A}}_S = \{\bar{a}_x| x \in \mathcal{S}\} $}\\
        
        \For{$E_2$ episodes}{
            Draw $N$ samples ($\mathcal{S}^\prime$) from $\mathcal{D}_b$\\ 
            Obtain unknown attributes for $\mathcal{S}^\prime$ using $f_{u}$: {$\bar{\mathcal{A}}_{S^\prime} = \{\bar{a}_x| x \in \mathcal{S}^\prime\} $}\\
            Calculate the lower-bound: $l_{MI}(\theta_{f_{I}}) = \frac{1}{N} \sum_{i=1}^{N} f_{I}({\bar{\mathcal{A}}_S}^{i},{\hat{\mathcal{A}}_S}^{i}) - \log(\frac{1}{N} \sum_{i=1}^{N} e^{f_{I}({\bar{\mathcal{A}}_{S^\prime}}^{i},{\hat{\mathcal{A}}_S}^{i})}) $\\
            Update $\theta_{f_{I}}$ using $Adam(-l_{MI}(\theta_{f_{I}}))$\\
        }
        
        Draw $N$ samples ($\mathcal{S}^{\prime\prime}$) from $\mathcal{D}_b$\\
        Obtain unknown attributes for $S^{\prime\prime}$ using $f_{u}$: {$\bar{A}_{S^{\prime\prime}} = \{\bar{a}_x| x \in S^{\prime\prime\}} $}\\
        Calculate loss: $l_{MI}(\theta_{f_{u}}) = \frac{1}{N} \sum_{i=1}^{N} f_{I}({\bar{\mathcal{A}}_S}^{i},{\hat{\mathcal{A}}_S}^{i}) - \log(\frac{1}{N} \sum_{i=1}^{N} e^{f_{I}({\bar{\mathcal{A}}_{S^{\prime\prime}}}^{i},{\hat{\mathcal{A}}_S}^{i})}) $\\
        Update $\theta_{f_{u}}$ using $Adam(l_{cls} +\lambda l_{MI}(\theta_{f_{u}}))$\\
    }

\end{algorithm*}

\section{Details of Learning Unknown Attributes and Training $g_{u}$}
\label{app_unk}

\begin{figure}
  \centering 
  \includegraphics[width=\linewidth]{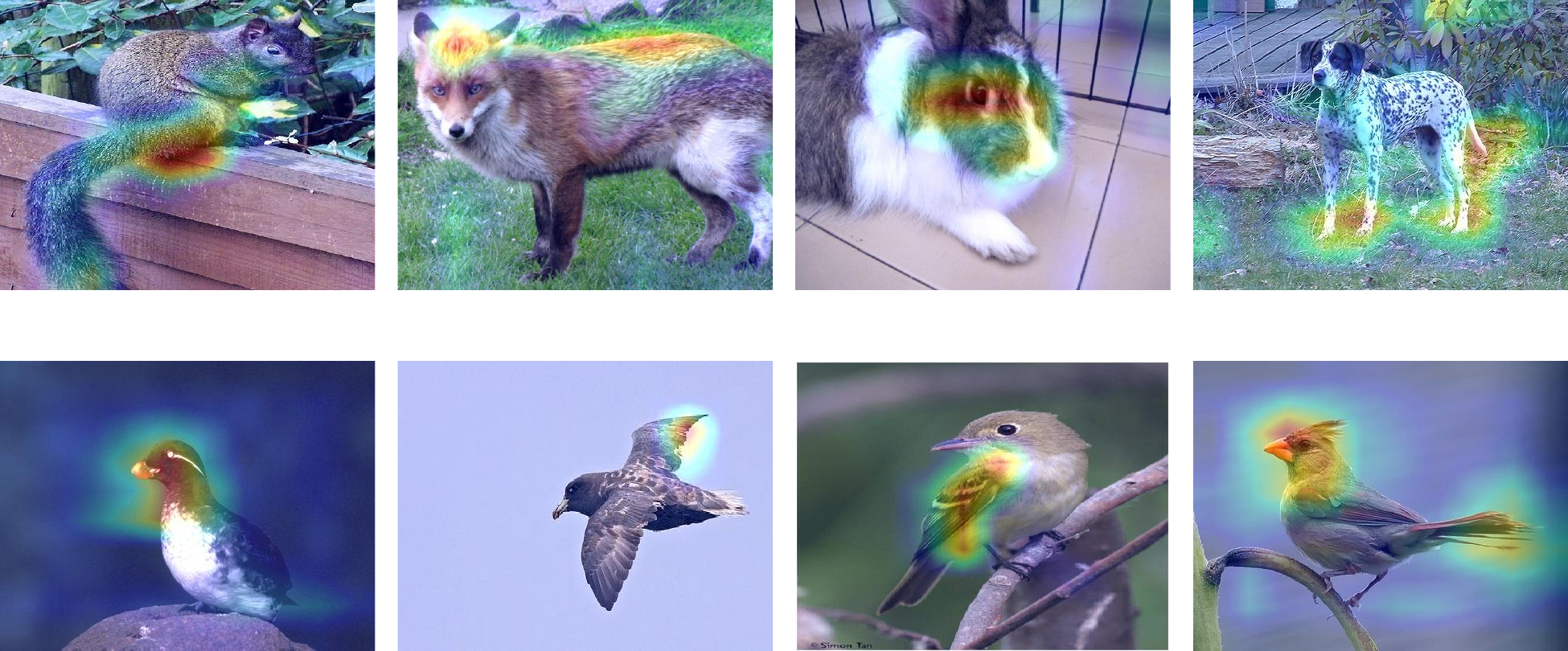}
  \caption{The Grad-cam visualization of four different unknown attributes on AWA dataset (top) and CUB dataset (bottom).}
  \label{fig.grad}
\end{figure}

To learn unknown attributes for classes, we use the same structure as $f_{h}$ for $f_{u}$. $f_I$ is an MLP with two hidden layers, both with a dimension of 50 and ReLU activation function. The input is the concatenation of human-friendly attributes and the output of $f_{u}$ that is used as unknown attributes, and the output is used to estimate mutual information using Equation \ref{eq__neural_mutual}. The networks $f_{u}$ and $f_{I}$ are trained simultaneously using the algorithm shown in Algorithm \ref{algo1}. In this algorithm, $E_1$ and $E_2$ are set to 200,000 and 10, respectively. Furthermore, $\lambda$ is to 2. The network is trained using Adam optimizer with a learning rate equal to $10^{-2}$, evaluated on 600 episodes sampled from validation classes every 500 episodes. The network with highest performance on validation episodes is used to extract embedding features acting as unknown attributes for each sample. Examples of Grad-Cam visualization for different unknown attributes of CUB and AWA datasets are shown in Figure \ref{fig.grad}.

To train $g_{u}$, the weights of our trained framework including $f_{h}$ and $g_{h}$ as well as $f_{u}$ are frozen and we just train $g_{u}$ which has a structure similar to the attribute selector network $g_{h}$ with the difference that this network will decide on the participation of the unknown attributes in each episode. Therefore, its output will be a scalar value. Similar to $g_{h}$, the input of $g_{u}$ is the prototypes of all classes participated in current episode with the difference that each prototype is calculated in the mixed space of unknown attributes and human-friendly attributes.

For each sample, the human-friendly attributes and unknown attributes are stacked to construct this mixed space. Then, the binary weight vector returned by $g_{h}$ is used as the weights of human-friendly attributes. Similarly, the output scalar value of $g_{u}$ is used as the weight for each of the unknown attributes.

\section{Evaluating the Impact of Prototype Permutations}
\label{app_lstm}
The attribute selector network $g_h$ in our model consists of a Bi-LSTM followed by a feed-forward network. The Bi-LSTM effectively aggregates the information from the attribute prototypes of all classes and provides unified features for them. However, the Bi-LSTM treats its input as a sequence with a specific order while the prototypes lack such an inherent order. In this section, we investigate how different permutations of the prototypes as Bi-LSTM input affect the few-shot classification accuracy. We sample 600 episodes from novel classes and assess the model's average accuracy across these episodes with varied permutations of classes as input to $g_h$. Within each episode, we randomly shuffle the classes and present their prototypes in this new order to $g_h$. We perform the few-shot classification on the same 600 episodes with random permutations for five times and calculate the average accuracy over the 600 episodes for each of these five evaluations. Then, the average over the set of accuracy evaluations with 95\% confidence interval are reported. Note that we did not use unknown attributes in this experiment. Furthermore, the L1 norm is not applied to the output of $g_{h}$ (i.e. $\eta = 0$). We perform the experiments with 5-way 1-shot, 5-way 3-shot and 5-way 5-shot settings on CUB and AWA datasets. The results presented in Table \ref{tab_permute_lstm}, reveal confidence intervals close to zero in all cases, indicating the model's insensitivity to the order of prototypes fed to the Bi-LSTM. This insensitivity arises from the episodic training approach, where the model encounters varying class orders, thus minimizing sensitivity to different permutations.

\begin{table}
\caption{The average performance of the model across five sets of evaluations on 600 few-shot episodes, considering various permutations of classes.}
\label{tab_permute_lstm}
\centering

\begin{tabular}{lccc}
\hline

\multirow{2}{*}{\textbf{Dataset}} & \multicolumn{2}{c}{\textbf{Setting}} \\
 & 1-shot & 3-shot & 5-shot\\
\hline

\textbf{CUB} & 56.26 $\pm$ 0.05 & 66.48 $\pm$ 0.02 & 71.41 $\pm$ 0.02\\
\textbf{AWA} & 42.67 $\pm$ 0.04 & 52.81 $\pm$ 0.02 & 56.1 $\pm$ 0.03\\

\end{tabular}

\end{table}

\section{Attribute predictor network performance}
\label{app_pre}

In this section, we evaluate the accuracy of the attribute predictor network. For each sample, the predicted attribute vector $s$ is compared with the ground-truth and an attribute accuracy is calculated for each sample. The mean and standard deviation of the attribute accuracy of individual samples in novel classes are reported in the last column of Table \ref{tab_3}.
Since the number of attributes present in each sample is usually much less than the number of absent attributes, we also report the mean accuracy of predicting present and absent attributes separately to assure that the model is not biased toward predicting the absence of the attributes due to the imbalanced nature of the attributes. The results are reported in the first two column of Table \ref{tab_3}. 
The results shows that the resulting models perform fairly well on predicting both present and absent attributes which demonstrates the effectiveness of the sample-wise weighted loss function in Eq. 2. Using the original CE loss instead of the sample-wise weighted loss function, almost all attributes were predicted as 0 (i.e. absent). 
For the aPY dataset, there is a 20\% gap between the accuracy of predicting absent and present attributes. This can be due to the weak relation between the base classes and novel classes in aPY so that the model has not been able to effectively transfer the knowledge learned on predicting present attributes of base classes to novel classes. In aPY, base classes are from Pascal and novel classes are from Yahoo classes as described in \ref{app_data}. 

We also evaluate the performance of the attribute predictor network on the attributes selected by the attribute selector network. For this experiment, we sample 600 episodes with a 5-way 5-shot setting from novel classes on each dataset and report the average performance in present, absent, and overall attribute prediction tasks for the attributes selected by the attribute selector network. The results are shown in Table \ref{tab_att_selected_1}. The observed results follow the same trend as in Table \ref{tab_3}, with a slight improvement in the performance of predicting present attributes. The trend was similar on 5-way 1-shot and 5-way 3-shot settings.

On average, the attribute predictor network is around 80\% accurate. This results demonstrates that less than perfect attribute predictors when incorporated into the proposed FSL method, can provide competitive FSL results. This can be attributed to the better generalization of higher level attributes to novel classes which is previously reported in \cite{cao2021concept} too.


\begin{table}
\caption{Performance of $f_{h}$ in predicting attributes of novel classes. The mean and standard deviation of the attribute prediction accuracy are reported. The accuracy of predicting absent attributes (AB), present attributes (PR) and the overall accuracy (OV) are reported separately.}
\label{tab_3}
\centering

\begin{tabular}{lccc}

\hline
\textbf{Dataset} & \textbf{AB} & \textbf{PR} &\textbf{OV}\\
\hline
\textbf{CUB} & 76.7 $\pm$ 0.7 & 78.8 $\pm$ 1.5 & 76.9 $\pm$ 0.7\\
\textbf{aPY} & 85.9 $\pm$ 0.6 & 62.2 $\pm$ 2.4 & 82.9 $\pm$ 0.6\\

\textbf{SUN} & 85.2 $\pm$ 0.5 & 80.1 $\pm$ 1.9 & 84.9 $\pm$ 0.6\\
\textbf{AWA} & 77.4 $\pm$ 0.6 & 71.4 $\pm$ 1.1 & 74.9 $\pm$ 0.7\\

\end{tabular}

\end{table}

\begin{table}
\caption{Performance of $f_{h}$ in predicting attributes selected by the attribute selector $g_h$ in 600 episodes of 5-way 5-shot classification sampled from novel classes. The mean and standard deviation of the attribute prediction accuracy are reported. The accuracy of predicting absent attributes (AB), present attributes (PR) and the overall accuracy (OV) are reported separately.}
\label{tab_att_selected_1}
\centering

\begin{tabular}{lccc}
\hline
\textbf{Dataset} & \textbf{AB} & \textbf{PR} &\textbf{OV}\\
\hline
\textbf{CUB} &  76.6$\pm$0.8 &  79$\pm$1.7  &  76.9$\pm$0.8 \\
\textbf{aPY} &  86.2$\pm$0.9  &  62.4$\pm$2.5 &  83.1$\pm$0.9 \\

\textbf{SUN} &  85.3$\pm$0.5  &  80.2$\pm$2.1  &  85$\pm$0.7 \\
\textbf{AWA} & 77.3$\pm$0.7  &  71.9$\pm$1.7  &  74.8$\pm$0.9 \\

\end{tabular}

\end{table}

\section{Performance Upperbound}
We report the performance of our model using true attributes as an upperbount in Table \ref{tab_upperbound}. Mean accuracy with 95\% confidence interval over 600 episodes sampled from novel classes is reported. We did not use unknown attributes in this experiment. Furthermore, the L1 norm is not applied to the output of $g_{h}$ (i.e. $\eta = 0$). As it can be seen, by improving the attribute predictor network, it is possible to obtain a much higher performance showing the effectiveness of our general framework and making decision based on using human-friendly attributes.   

\begin{table}
\caption{The accuracy of the model using true attributes as an upperbound. Mean accuracy with 95\% confidence interval over 600 episodes sampled from novel classes is reported.}
\label{tab_upperbound}
\centering

\begin{tabular}{lccc}
\hline

\multirow{2}{*}{\textbf{Dataset}} & \multicolumn{3}{c}{\textbf{Setting}} \\
 & 1-shot & 3-shot & 5-shot\\
\hline

\textbf{CUB} &  61.1 $\pm$ 1.1& 77.1 $\pm$ 0.7 & 82.8 $\pm$ 0.6\\
\textbf{aPY} & 67.3 $\pm$ 1 & 79.4 $\pm$ 0.8 & 80.8 $\pm$ 0.8\\
\textbf{SUN} & 60.1 $\pm$ 1 & 75.8 $\pm$ 0.8 & 78.9 $\pm$ 0.8\\
\textbf{AWA} & 85.9 $\pm$ 0.9 & 94.2 $\pm$ 0.5 & 100\\

\end{tabular}

\end{table}

\section{Quality of the selected attributes using the online attribute selection mechanism}
\label{app_selected_attributes}
\begin{figure}
  \centering 
  \includegraphics[width=\linewidth]{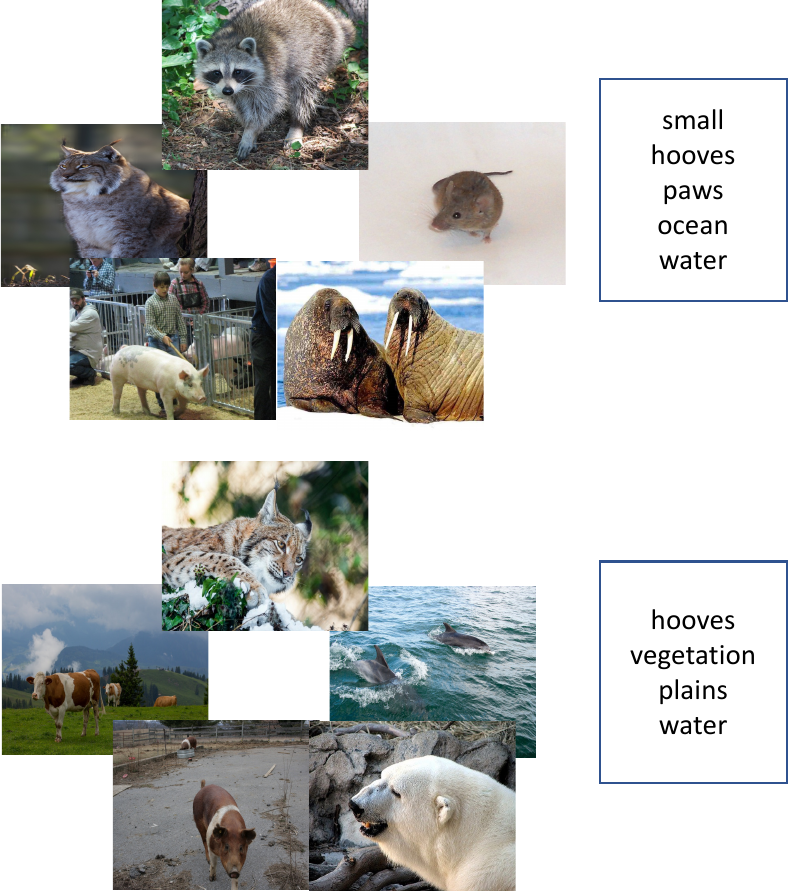}
 \caption{Qualitative examples: The attributes selected by online attribute selection mechanism for two 5-way 1-shot episodes on AWA dataset when $\gamma=0.01$. The selected attributes very well address the differences between images.}
  \label{fig.atts}
\end{figure}

In Figure \ref{fig.atts}, we have shown the attributes selected in two real episodes with 5-way 1-shot setting on AWA dataset.  In this experiment, to better evaluate the quality of the selected attributes, we have increased the loss term $l_1$ and set its weight to 0.01. It can be seen that the selected attributes very well relevant and address the differences between images, implying the effectiveness of the online attribute selection mechanism.

\section{Human Intervention}
\label{app_intervention}

We do intervention on the framework performing in joint attributes space (with unknown attributes) and when only human-friendly attributes are used (without unknown attributes). 600 episodes are sampled from novel classes and average accuracy before and after intervention with two intervention ratio (5\% and 10\%) are reported. It should be noted that in this experiment, unknown attributes are employed in all 600 episodes when evaluating intervention in joint attributes space. The results of this experiment are presented in Table \ref{tab_11}.

\begin{table*}
\caption{Intervention on AWA and CUB in 5-way 1-shot setting. We report the performance with respect to intervening 5\% ($r=5\%$) and 10\% ($r=10\%$) of human-friendly attributes selected by $g_{h}$ in each episode. Intervention is performed on misclassified query samples. The performance of the model without any intervention ($r=0\%$) for each setting is also reported for comparison.} 
\label{tab_11}

\centering
\begin{tabular}{lccc|ccc}
\hline
\multirow{2}{*}{\textbf{Setting}} & \multicolumn{3}{c|}{\textbf{CUB}} &  
\multicolumn{3}{c}{\textbf{AWA}} \\
  & $r=0\%$ & $r=5\%$ & $r=10\%$ & $r=0\%$ & $r=5\%$ & $r=10\%$ \\

\hline

With unknown attributes & 63.9 & 65.6 & 66.7 & 46.4 & 47.8 & 48.9 \\ 

Without unknown attributes & 56.3 & 59 & 60.3 & 42.6 & 45.2 & 46.7 \\

\hline

\end{tabular}
\end{table*}

\bibliographystyle{elsarticle-num}
\bibliography{references}

\end{document}